\documentclass[a4paper,letter]{article}

\usepackage{amsmath}
\usepackage{amsthm}
\usepackage{graphicx}
\usepackage{times}
\usepackage{subcaption}

\usepackage{natbib}
\setcitestyle{authoryear,round,citesep={;},aysep={,},yysep={;}}

\usepackage[margin=1in]{geometry}

\usepackage{amsmath,amsfonts,bm}

\def\eqref#1{equation~\ref{#1}}
\def\ceil#1{\lceil #1 \rceil}

\def\1{\bm{1}}

\def\eps{{\epsilon}}

\DeclareMathAlphabet{\mathsfit}{\encodingdefault}{\sfdefault}{m}{sl}
\SetMathAlphabet{\mathsfit}{bold}{\encodingdefault}{\sfdefault}{bx}{n}

\newcommand{\E}{\mathbb{E}}

\newcommand{\R}{\mathbb{R}}

\usepackage{hyperref}
\usepackage{url}

\def\rch{\operatorname{rch}} % reach of a manifold
\def\vol{\operatorname{vol}} % volume
\def\nn{\operatorname{nn}} % nearest neighbor
\def\Vor{\operatorname{Vor}}

\newtheorem{theorem}{Theorem} % section

\newtheorem{defin}{Definition}

\title{Adversarial Training with Voronoi Constraints}

\author{Marc Khoury \footnote{khoury@eecs.berkeley.edu}\\
University of California, Berkeley
\and
Dylan Hadfield-Menell \footnote{dhm@eecs.berkeley.edu}\\
University of California, Berkeley
}

\begin{document}

\maketitle

\begin{abstract}
Adversarial examples are a pervasive phenomenon of machine learning models where seemingly imperceptible perturbations to the input lead to misclassifications for otherwise statistically accurate models. We propose a geometric framework, drawing on tools from the manifold reconstruction literature, to analyze the high-dimensional geometry of adversarial examples. In particular, we highlight the importance of codimension: for low-dimensional data manifolds embedded in high-dimensional space there are many directions off the manifold in which an adversary could construct adversarial examples. Adversarial examples are a natural consequence of learning a decision boundary that classifies the low-dimensional data manifold well, but classifies points near the manifold incorrectly. Using our geometric framework we prove that adversarial training is sample inefficient, and show sufficient sampling conditions under which nearest neighbor classifiers and ball-based adversarial training are robust. Finally we introduce adversarial training with Voronoi constraints, which replaces the norm ball constraint with the Voronoi cell for each point in the training set. We show that adversarial training with Voronoi constraints produces robust models which significantly improve over the state-of-the-art on MNIST and are competitive on CIFAR-10.

\textbf{keywords:} adversarial examples, adversarial training, high-dimensional geometry, Voronoi diagrams, robustness, generalization

\end{abstract}

\thispagestyle{empty}
\setcounter{page}{0}
\newpage

\section{Introduction}
\label{sec:intro}
Deep learning at scale has led to breakthroughs on important problems
in computer vision~(\cite{Krizhevsky12}), natural language processing~(\cite{Wu16}),
and robotics~(\cite{Levine15}). Shortly thereafter, the
interesting phenomena of \emph{adversarial examples} was observed. A seemingly
ubiquitous property of machine learning models where perturbations of
the input that are imperceptible to humans reliably lead to confident
incorrect classifications (\cite{Szegedy13,Goodfellow14}). 
%There has been substantial subsequent work to understand this apparent brittleness and provide methods to train models that are robust to such perturbations.
What has ensued is a standard story from the security literature: a
game of cat and mouse where defenses are proposed only to be quickly
defeated by stronger attacks (\cite{Athalye18}). This has led
researchers to develop methods which are provably robust under
specific attack models (\cite{Wong18a, Sinha18, Raghunathan18, Mirman18}) as well as emperically strong heuristics (\cite{Madry17}). As machine
learning proliferates into society, including security-critical
settings like health care~(\cite{Esteva17}) or autonomous
vehicles~(\cite{Codevilla18}), it is crucial to develop methods that
allow us to understand the vulnerability of our models and design
appropriate counter-measures.

In this paper, we propose a geometric framework for analyzing the
phenomenon of adversarial examples. We leverage the observation that
datasets encountered in practice exhibit low-dimensional structure
despite being embedded in very high-dimensional input spaces. This
property is colloquially referred to as the ``Manifold Hypothesis'':
the idea that low-dimensional structure of `real' data leads to
tractable learning. We model data as being sampled from class-specific
low-dimensional manifolds embedded in a high-dimensional space. We consider a threat model where an adversary
may choose \emph{any} point on the data manifold to perturb by
$\epsilon$ in order to fool a classifier. In order to be robust to
such an adversary, a classifier must be correct everywhere in an
$\epsilon$-tube around the data manifold. Observe that, even though
the data manifold is a low-dimensional object, this tube has the same
dimension as the entire space the manifold is embedded in. Our
analysis argues that adversarial examples are a natural consequence of learning a decision boundary
that classifies all points on a low-dimensional data manifold correctly, but
classifies many points near the manifold incorrectly.
The high \emph{codimension}, the difference between
the dimension of the data manifold and the dimension of the embedding
space, is a key source of the pervasiveness of adversarial examples.

Our paper makes the following contributions. First, we develop a geometric framework,
inspired by the manifold reconstruction literature, that formalizes
the manifold hypothesis described above and our attack model. 
Second, we highlight the role \emph{codimension}
plays in vulnerability to adversarial examples. As the codimension increases,
there are an increasing number of directions off the data manifold in which to construct
adversarial perturbations. Prior work has attributed vulnerability
to adversarial examples to input dimension (\cite{Gilmer18, Shafahi19}). 
%This is the first work that investigates the
%role of \emph{codimension} in adversarial examples. Interestingly, we find that different 
%classification algorithms are less sensitive to changes in codimension.
Third, we apply this framework to analyze the standard approach to adversarial training.
We define a theoretical model $\mathcal{L}$ of adversarial training (see Definition \ref{def:advtrain}),
which guarantees correctness in the $\|\cdot\|_{p}$-balls centered on training data,
and prove that $\mathcal{L}$ is insufficient to learn robust decision boundaries with
realistic amounts of data. We show that nearest neighbor classifiers do not suffer from
this insufficiency, due to geometric properties of their
decision boundary away from data.
%and thus represent a potentially robust classification algorithm. 
Fourth we propose a modification to the standard 
paradigm of adversarial training. We replace the $\|\cdot\|_{p}$-ball constraint with 
the Voronoi cells of the training data, which have several advantages
detailed in Section~\ref{sec:advtrainvor}. In particular, we need not set the
maximum perturbation size $\epsilon$ as part of the training procedure.
%; the Voronoi cells adapt to the local feature size of the training set. 
In Section~\ref{sec:experiments} we show that adversarial
training with Voronoi constraints gives state-of-the-art robustness results on MNIST and competitive results on CIFAR-10.

\section{Related Work}
\subsection{Adversarial Examples}
Some previous work has considered the relationships between adversarial examples and high dimensional geometry. \cite{Franceschi18} explore the robustness of classifiers to random noise in terms of distance to the decision boundary, under the assumption that the decision boundary is locally flat. The work of \cite{Gilmer18} experimentally evaluated the setting of two concentric under-sampled $499$-spheres embedded in $\R^{500}$, and concluded that adversarial examples occur on the data manifold. In contrast, we present a geometric framework for proving robustness guarantees for learning algorithms, that makes no assumptions on the decision boundary. We carefully sample the data manifold in order to highlight the importance of \emph{codimension}; adversarial examples exist \emph{even} when the manifold is perfectly classified. Additionally we explore the importance of the spacing between the constituent data manifolds, sampling requirements for learning algorithms, and the relationship between model complexity and robustness. 

\cite{Wang18} explore the robustness of $k$-nearest neighbor classifiers to adversarial examples. In the setting where the Bayes optimal classifier is uncertain about the true label of each point, they show that $k$-nearest neighbors is not robust if $k$ is a small constant. They also show that if $k \in \Omega(\sqrt{dn\log{n}})$, then $k$-nearest neighbors is robust. Using our geometric framework we show a complementary result: in the setting where each point is certain of its label, $1$-nearest neighbors is robust to adversarial examples.

The decision and medial axes defined in Section~\ref{sec:geom} are maximum margin decision boundaries. Hard margin SVMs define define a linear separator with maximum margin, maximum distance from the training data (\cite{Cortes95}). Kernel methods allow for maximum margin decision boundaries that are non-linear by using additional features to project the data into a higher-dimensional feature space (\cite{Taylor04}). The decision and medial axes generalize the notion of maximum margin to account for the arbitrary curvature of the data manifolds. There have been attempts to incorporate maximum margins into deep learning (\cite{Sun16,Liu16,Liang17,Elsayed18}), often by designing loss functions that encourage large margins at either the output (\cite{Sun16}) or at any layer (\cite{Elsayed18}). In contrast, the decision axis is defined on the input space and we use it as an analysis tool for proving guarantees. 

\subsection{Manifold Reconstruction}
Manifold reconstruction is the problem of discovering the structure of a $k$-dimensional manifold embedded in $\R^d$, given \emph{only} a set of points sampled from the manifold. A large vein of research in manifold reconstruction develops algorithms that are \emph{provably good}: if the points sampled from the underlying manifold are sufficiently dense, these algorithms are guaranteed to produce a geometrically accurate representation of the unknown manifold with the correct topology. The output of these algorithms is often a \emph{simplicial complex}, a set of simplices such as triangles, tetrahedra, and higher-dimensional variants, that approximate the unknown manifold. In particular these algorithms output subsets of the Delaunay triangulation, which along with their geometric dual the Voronoi diagram, have properties that aid in proving geometric and topological guarantees (\cite{Edelsbrunner97}).

The field first focused on curve reconstruction in $\R^2$ (\cite{Amenta98}) and subsequently in $\R^3$ (\cite{Dey99}). Soon after algorithms were developed for surface reconstruction in $\R^3$, both in the noise-free setting (\cite{Amenta99, Amenta02}) and in the presence of noise (\cite{Dey04}). We borrow heavily from the analysis tools of these early works, including the medial axis and the reach. However we emphasize that we have adapted these tools to the learning setting. To the best of our knowledge, our work is the first to consider the medial axis under different norms.

In higher-dimensional embedding spaces (large $d$), manifold reconstruction algorithms face the \emph{curse of dimensionality}. In particular, the Delaunay triangulation, which forms the bedrock of algorithms in low-dimensions, of $n$ vertices in $\R^d$ can have up to $\Theta(n^{\ceil{d/2}})$ simplices. To circumvent the curse of dimensionality, algorithms were proposed that compute subsets of the Delaunay triangulation restricted to the $k$-dimensional tangent spaces of the manifold at each sample point (\cite{Boissonnat14}). Unfortunately, progress on higher-dimensional manifolds has been limited due to the presence of so-called ``sliver'' simplices, poorly shaped simplices that cause in-consistences between the local triangulations constructed in each tangent space (\cite{Cheng05, Boissonnat14}). Techniques that provably remove sliver simplices have prohibitive sampling requirements (\cite{Cheng00, Boissonnat14}). Even in the special case of surfaces ($k=2$) embedded in high dimensions ($d > 3$), algorithms with practical sampling requirements have only recently been proposed (\cite{Khoury16}). Our use of tubular neighborhoods as a tool for analysis is borrowed from \cite{Dey05} and \cite{Khoury16}. 

In this paper we are interested in \emph{learning} robust decision boundaries, \emph{not} reconstructing the underlying data manifolds, and so we avoid the use of Delaunay triangulations and their difficulties entirely. In Section~\ref{sec:proverobust} we present robustness guarantees for two learning algorithms in terms of a sampling condition on the underlying manifold. These sampling requirements scale with the dimension of the underlying manifold $k$, \emph{not} with the dimension of the embedding space $d$.

\section{The Geometry of Data}
\label{sec:geom}
We model data as being sampled from a set of low-dimensional manifolds
(with or without boundary) embedded in a high-dimensional space $\R^d$.
We use $k$ to denote the dimension of a manifold $\mathcal{M} \subset \R^d$.
The special case of a $1$-manifold is called a \emph{curve}, and
a $2$-manifold is a \emph{surface}.
The \emph{codimension} of $\mathcal{M}$ is $d - k$, the difference between
the dimension of the manifold and the dimension of the embedding space.
The ``Manifold Hypothesis'' is the observation that in practice, data is often
sampled from manifolds, usually of high codimension.

In this paper we are primarily interested in the classification problem. Thus we model data as being sampled from $C$ \emph{class manifolds} $\mathcal{M}_{1}, \ldots, \mathcal{M}_{C}$, one for each class. When we wish to refer to the entire space from which a dataset is sampled, we refer to the $\emph{data manifold}$ $\mathcal{M} = \cup_{1 \leq j \leq C} \mathcal{M}_j$. We often work with a finite sample of $n$ points, $X \subset \mathcal{M}$, and we write $X = \{ X_1, X_2, \ldots, X_n \}$. Each sample point $X_i$ has an accompanying class label $y_i \in \{ 1, 2, \ldots, C \}$ indicating which manifold $\mathcal{M}_{y_i}$ the point $X_i$ is sampled from.

Consider a $\|\cdot\|_{p}$-ball $B$ centered at some point $c \in \R^d$ and imagine growing $B$ by increasing its radius starting from zero. For nearly all starting points $c$, the ball $B$ eventually intersects one, \emph{and only one}, of the $\mathcal{M}_i$'s. Thus the nearest point to $c$ on $\mathcal{M}$, in the norm $\|\cdot\|_{p}$, lies on $\mathcal{M}_{i}$. 
%(Note that the nearest point on $\mathcal{M}_{i}$ need not be unique.) 

\begin{figure}[h!]
\begin{center}
\includegraphics[width=0.85\textwidth]{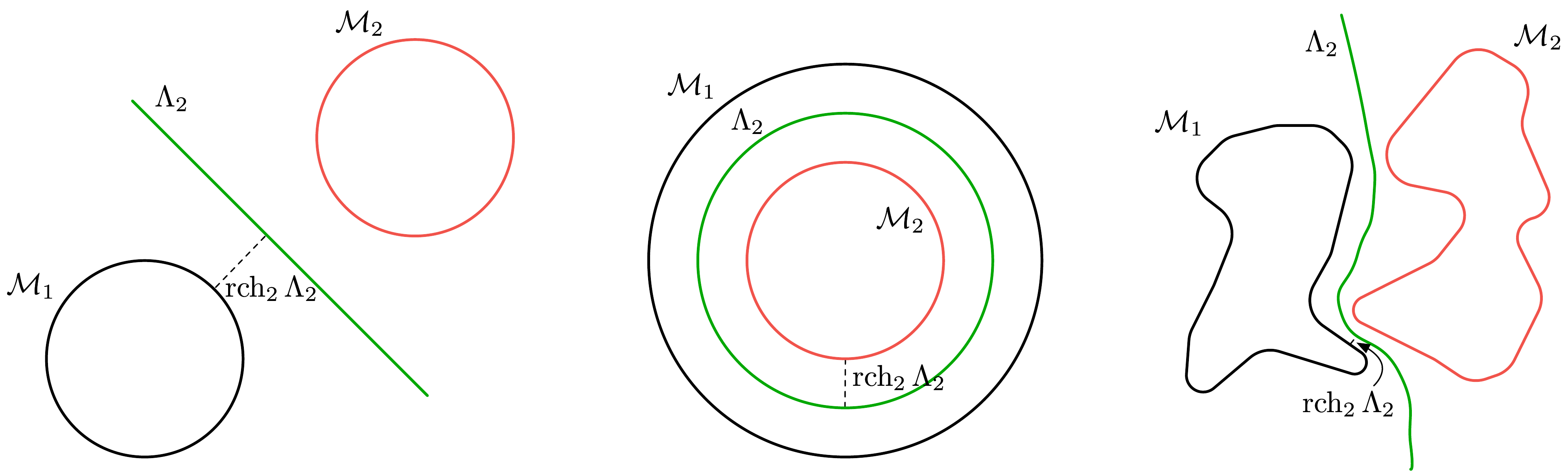}
\caption{Examples of the decision axis $\Lambda_2$, shown here in green, for different data manifolds. Intuitively, the decision axis captures an optimal decision boundary between the data manifolds. It's optimal in the sense that each point on the decision axis is as far away from each data manifold as possible. Notice that in the first example, the decision axis coincides with the maximum margin line.}
\label{fig:medialaxis}
\end{center}
\end{figure}

The \emph{decision axis} $\Lambda_{p}$ of $\mathcal{M}$ is the set of points $c$ such that the boundary of $B$ intersects two or more of the $\mathcal{M}_{i}$, but the interior of $B$ does not intersect $\mathcal{M}$ at all. In other words, the decision axis $\Lambda_p$ is the set of points that have two or more closest points, in the norm $\|\cdot\|_{p}$, \emph{on distinct class manifolds}. See Figure \ref{fig:medialaxis}. The decision axis is inspired by the medial axis, which was first proposed by \cite{Blum67} in the context of image analysis and subsequently modified for the purposes of curve and surface reconstruction by \cite{Amenta98, Amenta02}. We have modified the definition to account for multiple class manifolds and have renamed our variant in order to avoid confusion in the future.

The decision axis $\Lambda_p$ can intuitively be thought of as a decision boundary that is optimal in the following sense. First, $\Lambda_p$ separates the class manifolds when they do not intersect. Second, each point of $\Lambda_p$ is as far away from the class manifolds as possible in the norm $\|\cdot\|_{p}$. As shown in the leftmost example in Figure \ref{fig:medialaxis}, in the case of two linearly separable circles of equal radius, the decision axis $\Lambda_{2}$ is exactly the line that separates the data with maximum margin. For arbitrary manifolds, $\Lambda_p$ generalizes the notion of maximum margin to account for the curvature of the class manifolds. 

Let $T \subset \R^d$ be any set. The \emph{reach} $\rch_{p}{(T; \mathcal{M})}$ of $\mathcal{M}$ is defined as $\inf_{x \in \mathcal{M}, y \in T} \|x - y\|_{p}$. When $\mathcal{M}$ is compact, the reach is achieved by the point on $\mathcal{M}$ that is closest to $T$ under the $\|\cdot\|_{p}$ norm. We will drop $\mathcal{M}$ from the notation when it is understood from context. 

Finally, an $\epsilon$-\emph{tubular neighborhood} of $\mathcal{M}$ is defined as $\mathcal{M}^{\epsilon, p} = \{x \in \R^d: \inf_{y \in \mathcal{M}}\|x - y\|_p \leq \epsilon\}$. That is, $\mathcal{M}^{\epsilon, p}$ is the set of all points whose distance to $\mathcal{M}$ under the metric induced by $\|\cdot\|_{p}$ is less than $\epsilon$. Note that while $\mathcal{M}$ is $k$-dimensional, $\mathcal{M}^{\epsilon, p}$ is always $d$-dimensional. Tubular neighborhoods are how we rigorously define adversarial examples. Consider a classifier $f: \R^d \rightarrow [C]$ for $\mathcal{M}$. An $\epsilon$-\emph{adversarial example} is a point $x \in \mathcal{M}_{i}^{\epsilon, p}$ such that $f(x) \neq i$. A classifier $f$ is robust to all $\epsilon$-adversarial examples when $f$ correctly classifies not only $\mathcal{M}$, but all of $\mathcal{M}^{\epsilon, p}$. 
%Thus the problem of being robust to adversarial examples is rightly seen as one of \emph{generalization}. 
In this paper we will be primarily concerned with exploring the conditions under which we can provably learn a decision boundary that correctly classifies $\mathcal{M}^{\epsilon, p}$. When $\epsilon < \rch_p{\Lambda_p}$, the decision axis $\Lambda_p$ is one decision boundary that correctly classifies $\mathcal{M}^{\epsilon, p}$. Throughout the remainder of the paper we will drop the $p$ in $\mathcal{M}^{\epsilon, p}$ from the notation, instead writing $\mathcal{M}^{\epsilon}$; the norm will always be clear from context.

%The geometric quantities defined above can be defined more generally for any distance metric $d(\cdot, \cdot)$. In this paper we will focus exclusively on the metrics induced by the norms $\|\cdot\|_{p}$ for $p > 0$. 

\section{Limitations of Adversarial Training}
\label{sec:proverobust}
Adversarial training, the process of training on adversarial examples generated in $\|\cdot\|_{p}$-balls around the training data, is a very natural approach to constructing robust models (\cite{Goodfellow14, Madry17}). In our notation this corresponds to training on samples drawn from $X^{\epsilon}$ for some $\epsilon$. Despite its simplicity, adversarial training has proven to be one of the most successful approaches to training robust deep networks. While natural, we show that there are simple settings where this approach is much less sample-efficient than other classification algorithms, if the \emph{only} guarantee is correctness in $X^{\epsilon}$.   

\begin{defin}[Adversarial Training]
\label{def:advtrain}
Let $X \subset \mathcal{M}$ be a finite training set. Define an adversarial training algorithm $\mathcal{L}$ as a learning algorithm that, given $X$, outputs a model $f_{\mathcal{L}}$ such that for every $x \in X$ with label $y$, and every $\hat{x} \in B(x, \rch_{p}{\Lambda_{p}})$, $f_{\mathcal{L}}(\hat{x}) = f_{\mathcal{L}}(x) = y$. Here $B(x, r)$ denotes the ball centered at $x$ of radius $r$ in the $\|\cdot\|_p$ norm.
\end{defin}

$\mathcal{L}$ is our \emph{theoretical model} of the standard approach to adversarial training (\cite{Goodfellow14, Madry17}). In words, $\mathcal{L}$ learns a model that outputs the same label for any $\|\cdot\|_{p}$-perturbation of $x$ up to $\rch_{p}{\Lambda_{p}}$ as it outputs for $x$. We will use $\mathcal{L}$ to analyze the limitations of the standard approach to adversarial training; in particular we will show that $\mathcal{L}$ is much less sample efficient at learning a robust decision boundary than other classification algorithms.

\begin{theorem}
\label{thm:classexists}
There exists a classification algorithm $\mathcal{A}$ that, for a particular choice of $\mathcal{M}$, correctly classifies $\mathcal{M}^{\epsilon}$ using exponentially fewer samples than are required for $\mathcal{L}$ to correctly classify $\mathcal{M}^{\epsilon}$. 
\end{theorem}

The reason for the sample inefficiency of $\mathcal{L}$ is the use of the $\|\cdot\|_{p}$-balls centered on the data to propagate the labels. As we will show below, the union of the balls around the data $X^{\epsilon}$ covers a vanishingly small fraction of $\mathcal{M}^{\epsilon}$ in high codimension settings. Thus the adversary is restricted to constructing adversarial examples in a negligible fraction of the neighborhood around the data manifold. In contrast, other algorithms, such as nearest neighbor classifiers, propagate labels using different geometric regions, such as the Voronoi cells which we will define in Section~\ref{sec:advtrainvor}. The \emph{main takeaway} of this paper is that the use of $\|\cdot\|_{p}$-balls centered on the data leads to sub-optimal results both in theory and, as we will show in Section~\ref{sec:experiments}, in practice.

Theorem~\ref{thm:classexists} follows from Theorems~\ref{thm:sampling} and \ref{thm:samplinggap}. In Theorems~\ref{thm:sampling} and \ref{thm:samplinggap} we will prove that a nearest neighbor classifier $f_{\nn}$ is one such classification algorithm. Nearest neighbor classifiers are naturally robust in high codimensions because the Voronoi cells of $X$ are \emph{elongated in the directions normal} to $\mathcal{M}$ when $X$ is dense (\cite{Dey07}).

Before we state Theorem \ref{thm:sampling} we must introduce a sampling condition on $\mathcal{M}$. A $\delta$-cover of a manifold $\mathcal{M}$ in the norm $\|\cdot \|_{p}$ is a finite set of points $X$ such that for every $x \in \mathcal{M}$ there exists $X_i$ such that $\|x - X_i\|_{p} \leq \delta$. Theorem~\ref{thm:sampling} gives a sufficient sampling condition for $f_{\mathcal{L}}$ to correctly classify $\mathcal{M}^{\epsilon}$ for all manifolds $\mathcal{M}$. Theorem~\ref{thm:sampling} also provides a sufficient sampling condition for a nearest neighbor classifier $f_{\nn}$ to correctly classify $\mathcal{M}^{\epsilon}$, which is substantially less dense than that of $f_{\mathcal{L}}$. Thus different classification algorithms have different sampling requirements in high codimensions.

\begin{theorem}
\label{thm:sampling}
Let $\mathcal{M} \subset \R^d$ be a $k$-dimensional manifold and let $\epsilon < \rch_p{\Lambda_p}$ for any $p > 0$. Let $f_{nn}$ be a nearest neighbor classifier and let $f_{\mathcal{L}}$ be the output of a learning algorithm $\mathcal{L}$ as described above. Let $X_{\nn}, X_{\mathcal{L}} \subset \mathcal{M}$ denote the training sets for $f_{\nn}$ and $\mathcal{L}$ respectively. We have the following sampling guarantees: 

\begin{enumerate}
\item If $X_{\nn}$ is a $\delta$-cover for $\delta \leq 2 (\rch_p{\Lambda_p} - \epsilon)$ then $f_{\nn}$ correctly classifies $\mathcal{M}^{\epsilon}$.
\item If $X_{\mathcal{L}}$ is a $\delta$-cover for $\delta \leq \rch_p{\Lambda_p} - \epsilon$ then $f_{\mathcal{L}}$ correctly classifies $\mathcal{M}^{\epsilon}$.
\end{enumerate}
\end{theorem}

The bounds on $\delta$ in Theorem~\ref{thm:sampling} are sufficient, but they are not always necessary. There exist manifolds where the bounds in Theorem~\ref{thm:sampling} are pessimistic, and less dense samples corresponding to larger values of $\delta$ would suffice. 

Next we will show a setting where bounds on $\delta$ similar to those in Theorem~\ref{thm:sampling} are \emph{necessary}. In this setting, the difference of a factor of $2$ in $\delta$ between the sampling requirements of $f_{\nn}$ and $f_{\mathcal{L}}$ leads to an exponential gap between the sizes of $X_{\nn}$ and $X_{\mathcal{L}}$ necessary to achieve identical robustness.

Define $\Pi_1 = \{x \in \R^d: \ell \leq x_1, \ldots, x_k \leq \mu \text{ and } x_{k+1} = \ldots = x_d = 0\}$; that is $\Pi_1$ is a subset of the $x_1$-$\ldots$-$x_k$-plane bounded between the coordinates $[\ell, \mu]$. Similarly define $\Pi_2 = \{x \in \R^d: \ell \leq x_1, \ldots, x_k \leq \mu \text{ and } x_{k+1} = \ldots = x_{d-1} = 0 \text{ and } x_d = 2\}$. Note that $\Pi_2$ lies in the subspace $x_d = 2$; thus $\rch_{2}{\Lambda_{2}} = 1$, where $\Lambda_2$ is the decision axis of $\Pi = \Pi_1 \cup \Pi_2$. In the $\|\cdot\|_{2}$ norm we can show that the gap in Theorem~\ref{thm:sampling} is necessary for $\Pi = \Pi_1 \cup \Pi_2$. Furthermore the bounds we derive for $\delta$-covers for $\Pi$ for both $f_{\nn}$ and $f_{\mathcal{L}}$ are tight. Combined with well-known properties of covers, we get that the ratio $|X_{\mathcal{L}}|/|X_{\nn}|$ is exponential in $k$.

\begin{theorem}
\label{thm:samplinggap}
Let $\Pi = \Pi_{1} \cup \Pi_{2}$ as described above. Let $X_{\nn}, X_{\mathcal{L}} \subset \Pi$ be minimum training sets necessary to guarantee that $f_{\nn}$ and $f_{\mathcal{L}}$ correctly classify $\mathcal{M}^{\epsilon}$. Then we have that
\begin{equation}
\frac{|X_{\mathcal{L}}|}{|X_{\nn}|} \in \Omega\left(2^{k/2}\right)
\end{equation}
\end{theorem}

We have shown that both $\mathcal{L}$ and nearest neighbor classifiers learn robust decision boundaries when provided sufficiently dense samples of $\mathcal{M}$. However there are settings where nearest neighbors is exponentially more sample-efficient than $\mathcal{L}$ in achieving the same amount of robustness. 

To shed light on why the ball-based learning algorithm $\mathcal{L}$ is so much less sample-efficient than nearest neighbor classifiers, we show that the volume $\vol{X^{\epsilon}}$ is often a vanishingly small percentage of $\vol{\mathcal{M}^{\epsilon}}$. For our theoretical model $\mathcal{L}$ this means that a vanishingly small fraction of $\mathcal{M}^{\epsilon}$ is guaranteed to have the correct label, and in practice this means that the adversary in adversarial training does not have the freedom to generate adversarial examples in the entirety of $\mathcal{M}^{\epsilon}$. For the remainder of this section we will consider the $\|\cdot\|_{2}$ norm. 

\begin{theorem}
\label{thm:manifoldvolumelowerbound}
Let $\mathcal{M} \subset \R^d$ be a $k$-dimensional manifold embedded in $\R^d$ such that $\vol_{k}{\mathcal{M}} < \infty$. Let $X \subset \mathcal{M}$ be a finite set of points sampled from $\mathcal{M}$. Suppose that $\epsilon \leq \rch_{2}{\Xi}$ where $\Xi$ is the medial axis of $\mathcal{M}$, defined as in \cite{Dey07}. Then the percentage of $\mathcal{M}^{\epsilon}$ covered by $X^{\epsilon}$ is upper bounded by
\begin{equation}
\label{equ:manifoldvolumelowerbound}
\frac{\vol{X^\epsilon}}{\vol{\mathcal{M}^\epsilon}} \in \mathcal{O}\left(\left(\frac{2\pi}{d - k}\right)^{k/2} \frac{\epsilon^k}{\vol_{k}{\mathcal{M}}} |X| \right).
\end{equation}
As the codimension $(d - k) \rightarrow \infty$, Equation~\ref{equ:manifoldvolumelowerbound} approaches $0$, for any fixed $|X|$.
\end{theorem}

In high codimension, even moderate under-sampling of $\mathcal{M}$ leads to a significant loss of coverage of $\mathcal{M}^{\epsilon}$ because the volume of the union of balls centered at the samples shrinks faster than the volume of $\mathcal{M}^{\epsilon}$. Theorem~\ref{thm:manifoldvolumelowerbound} states that in high codimensions the fraction of $\mathcal{M}^{\epsilon}$ covered by $X^{\epsilon}$ goes to $0$. Almost nothing is covered by $X^{\epsilon}$ for training set sizes that are realistic in practice. Thus $X^{\epsilon}$ is a poor model of $\mathcal{M}^{\epsilon}$, and high classification accuracy on $X^{\epsilon}$ does not imply high accuracy in $\mathcal{M}^{\epsilon}$. 

%To give a sense of how rapidly the upper bound in Equation \ref{equ:manifoldvolumelowerbound} goes to $0$, we plot its value in Figure \ref{fig:planelowerbound} (Left) for the special case of $k$-dimensional planes $\Pi$. Figure~\ref{fig:planelowerbound} (Right) shows the number of samples that are sufficient for $X^{1}$ to cover $\Pi$ compared to a lower bound on the number of samples necessary to cover the $1$-tubular neighborhood $\Pi^{1}$. 

%\begin{figure*}
%\begin{center}
%\begin{subfigure}{0.4\textwidth}
%\includegraphics[width=0.98\linewidth]{figures/volume-plane-upper-bound}
%\end{subfigure}
%\begin{subfigure}{0.55\textwidth}
%\includegraphics[width=0.98\linewidth]{figures/plane-lower-bound}
%\end{subfigure}
%\caption{We plot the upper bound in Equation \ref{equ:manifoldvolumelowerbound} on the left for the special case of $\Pi$. As the codimension increases, the percentage of volume of $\Pi^{1}$ covered by $1$-balls around the $1$-sample approaches $0$. On the right we plot the number of samples sufficient to cover $\Pi$, shown in blue, against the number of samples necessary to cover $\Pi^{1}$, shown in orange, as the codimension increases.}
%\label{fig:planelowerbound}
%\end{center}
%\end{figure*}

Approaches that produce robust classifiers by generating adversarial examples in the $\epsilon$-balls centered on the training set do not accurately model $\mathcal{M}^{\epsilon}$, and it will take \emph{many} more samples to do so. If the method behaves arbitrarily outside of the $\epsilon$-balls that define $X^{\epsilon}$, adversarial examples will still exist and it will likely be easy to find them. The reason deep learning has performed so well on a variety of tasks, in spite of the brittleness made apparent by adversarial examples, is because it is much easier to perform well on $\mathcal{M}$ than it is to perform well on $\mathcal{M}^{\epsilon}$.  

\section{Adversarial Training with Voronoi Constraints}
\label{sec:advtrainvor}
\cite{Madry17} formalize adversarial training by introducing the robust objective 
\begin{equation}
\label{equ:robustopt}
\min_{\theta} \E_{(x,y)\in \mathcal{D}}\left[\max_{\hat{x} \in B(x, \epsilon)} L(\hat{x}, y; \theta)\right]
\end{equation}
 where $\mathcal{D}$ is the data distribution and $B$ is a $\|\cdot\|_{p}$-ball centered at $x$ with radius $\epsilon$. Their main contribution was the use of a strong adversary which used projected gradient descent to solve the inner optimization problem.

In Sections~\ref{sec:proverobust}, we showed that adversarial training formalized using the geometric constraint $B$ is sample inefficient, because the adversary is restricted to a negligible fraction of the $\epsilon$-tubular neighborhood around the data distribution. To remedy this we replace the $\|\cdot\|_p$-ball constraint with a different geometric constraint, namely the Voronoi cell at $x$. That is, we formalize the adversarial training objective as 
\begin{equation}
\label{equ:robustoptours}
\min_{\theta} \E_{(x,y)\in \mathcal{D}}\left[\max_{\hat{x} \in \Vor_p{x}}  L(\hat{x}, y; \theta)\right]
\end{equation}
where 
\begin{equation}
\label{equ:vorcell}
\Vor_{p}{x} = \{x' \in \R^d: \|x - x'\|_{p} \leq \|z - x'\|_{p} \; \forall z \in X \backslash \{x\}\}.
\end{equation}
In words, the Voronoi cell $\Vor_{p}{x}$ of $x$ is the set of all points in $\R^d$ that are closer to $x$ than to any other sample in $X$. 

The Voronoi cell constraint has many advantages over the $\|\cdot\|_{p}$-ball constraint. First the Voronoi cells \emph{partition} the entirety of $\R^{d}$ and so the interiors of Voronoi cells generated by samples from different classes do not intersect. This is in contrast to $\|\cdot\|_{p}$-balls which may intersect for sufficiently large $\epsilon$ and cause problems for optimization. In particular the Voronoi cells partition $\mathcal{M}^{\epsilon}$ and, for dense samples, are elongated in the directions normal to the data manifold. Thus the Voronoi cells are well suited for high codimension settings. Second, the size of the Voronoi cells adapts to the data distribution. A Voronoi cell generated by a sample which is close to samples from a different class manifold is smaller, while those further away are larger. Thus we do not need to set a value for $\epsilon$ in the optimization procedure. The constraint naturally adapts to the largest value of $\epsilon$ possible locally on the data manifold. Third, the Voronoi cells enjoy the sample efficiency of $f_{\nn}$ in Theorem~\ref{thm:sampling}, because the Voronoi cells define the nearest neighbor decision boundary. In summary, the Voronoi constraint gives the adversary the freedom to explore the entirety of the tubular neighborhood around $\mathcal{M}$. 

At each iteration we must solve the inner optimization problem  
\begin{equation}
\label{equ:advopt}
\begin{aligned}
& \underset{\hat{x}}{\text{maximize}}
& & L(\hat{x}, y; \theta) \\
& \text{subject to}
& & \|x - \hat{x}\|_{p} - \|z - \hat{x}\|_{p} \leq 0 \; \forall z \in X - \{x\}. 
\end{aligned}
\end{equation}

When $p = 2$ the Voronoi cells are convex and so we can project a point onto a Voronoi cell by solving a quadratic program. Thus we can solve Problem \ref{equ:advopt} using projected gradient descent, as in \cite{Madry17}. When $p \neq 2$ the Voronoi cells are not necessarily convex. In this setting there are many approaches, such as barrier and penalty methods, one might employ to approximately solve Problem \ref{equ:advopt} (\cite{Boyd04}). However we found that the following heuristic is both fast and works well in practice. At each iteration of the outer training loop, for each training sample $x$ in a batch, we generate adversarial examples by taking iterative steps in the direction of the gradient starting from $x$. However instead of projecting onto a constraint after each iterative step, we instead check if any of the Voronoi constraints of $x$ shown in Equation \ref{equ:vorcell} are violated. If no constraint is violated we perform the iterative update, otherwise we simply stop performing updates for $x$. 

Problem \ref{equ:advopt} has $n - 1$ constraints, one for each sample in $X \backslash \{x\}$. In practice however very few samples contribute to the Voronoi cell of $x$. At each iteration, we perform a nearest neighbor search query to find the $m$ nearest samples to $x$ in each other class. That is we search for $m(C-1)$ samples where $C$ is the number of classes. We do not impose constraints from samples in the same class as $x$; there is no benefit to restricting the adversary's movement with the tubular neighborhood around the class manifold of $x$. In our experiments we set $m = 10$.   

\section{Experiments}
\label{sec:exp}
\label{sec:experiments}

%\subsection{Setup}
\textbf{Datasets.} 
To investigate how the codimension of a dataset influences robustness we introduce two synthetic datasets, {\sc Circles} and {\sc Planes}, which allow us to carefully vary the codimension while maintaining dense samples. The {\sc Circles} dataset consists of two concentric circles in the $x_1$-$x_2$-plane, the first with radius $r_1 = 1$ and the second with radius $r_2 = 3$, so that $\rch_2{\Lambda_2} = 1$. We densely sample $1000$ random points on each circle for both the training and the test sets. The {\sc Planes} dataset consists of two $2$-dimensional planes, the first in the $x_d=0$ and the second in $x_d = 2$, so that $\rch_2{\Lambda_2} = 1$. The first two axis of both planes are bounded as $-10 \leq x_1, x_2 \leq 10$, while $x_3 = \ldots = x_{d-1} = 0$. We sample the training set at the vertices of a regular grid with side length $\sqrt{2}$, and the test set at the centers of the grid cubes. We also evaluate on MNIST and CIFAR-10.
%With this spacing, both planes are sampled so that the $1$-tubular neighborhood $X^{1}$ covers the underlying planes, where $X$ is the training set. 
%We evaluate our adversarial training approach on MNIST and CIFAR-10. 

\textbf{Models.}
Our controlled experiments on synthetic data consider a fully connected network with 1 hidden layer, 100 hidden units, and ReLU activations. We set the learning rate for Adam (\cite{Kingma15}) as $\alpha = 0.1$. Our experimental results are averaged over 20 retrainings. For a fair comparison, our experiments on MNIST and CIFAR-10 use the same model architectures as in \cite{Madry17}. 
%The model for MNIST consists of two convolutional layers with 32 and 64 filters respectively, each followed by $2 \times 2$ max pooling. After the two convolutional layers, there are two fully connected layers each with $1024$ hidden units. The model for CIFAR-10 is a resnet with five residual units. Additionally we performed data augmentation using random crops and flips. 
We train the MNIST model using Adam for 100 epochs and the CIFAR-10 model using SGD for 250 epochs.

\textbf{Attacks.}
We consider two attacks, the fast gradient sign method (FGSM) (\cite{Goodfellow14}) and the basic iterative method (BIM) (\cite{Kurakin16}). We use the implementations provided in the cleverhans library \cite{Papernot18}

\textbf{Accuracy measures.}
We plot the minimum classification accuracy across our suite of attacks as a function of $\epsilon$, for each of our datasets. Additionally we report the \emph{normalized area under the curve} (NAUC) defined as 
\begin{equation}
\operatorname{NAUC}(\operatorname{acc}) = \frac{1}{\epsilon_{\max}}\int_{0}^{\epsilon_{\max}} \operatorname{acc}(\epsilon) \; d\epsilon,
\end{equation}
where $\operatorname{acc}: [0, \epsilon_{\max}] \rightarrow [0, 1]$ measures the classification accuracy and $\eps_{\max}$ is the largest perturbation considered. Note that NAUC $\in [0,1]$ with higher values corresponding to more robust models. 

\subsection{High Codimension Reduces Robustness}
\label{ssec:codim}
Section~\ref{sec:proverobust} suggests that as the codimension increases it should become easier to find adversarial examples, which Figure \ref{fig:codimexp} (Top Left) shows on the {\sc Circles} dataset. We see a steady decrease in robustness as we increase the codimension.

\subsection{Adversarial Training in High Codimensions}
\begin{figure}[h!]
\begin{center}
\begin{subfigure}{0.46\textwidth}
\includegraphics[width=0.99\linewidth]{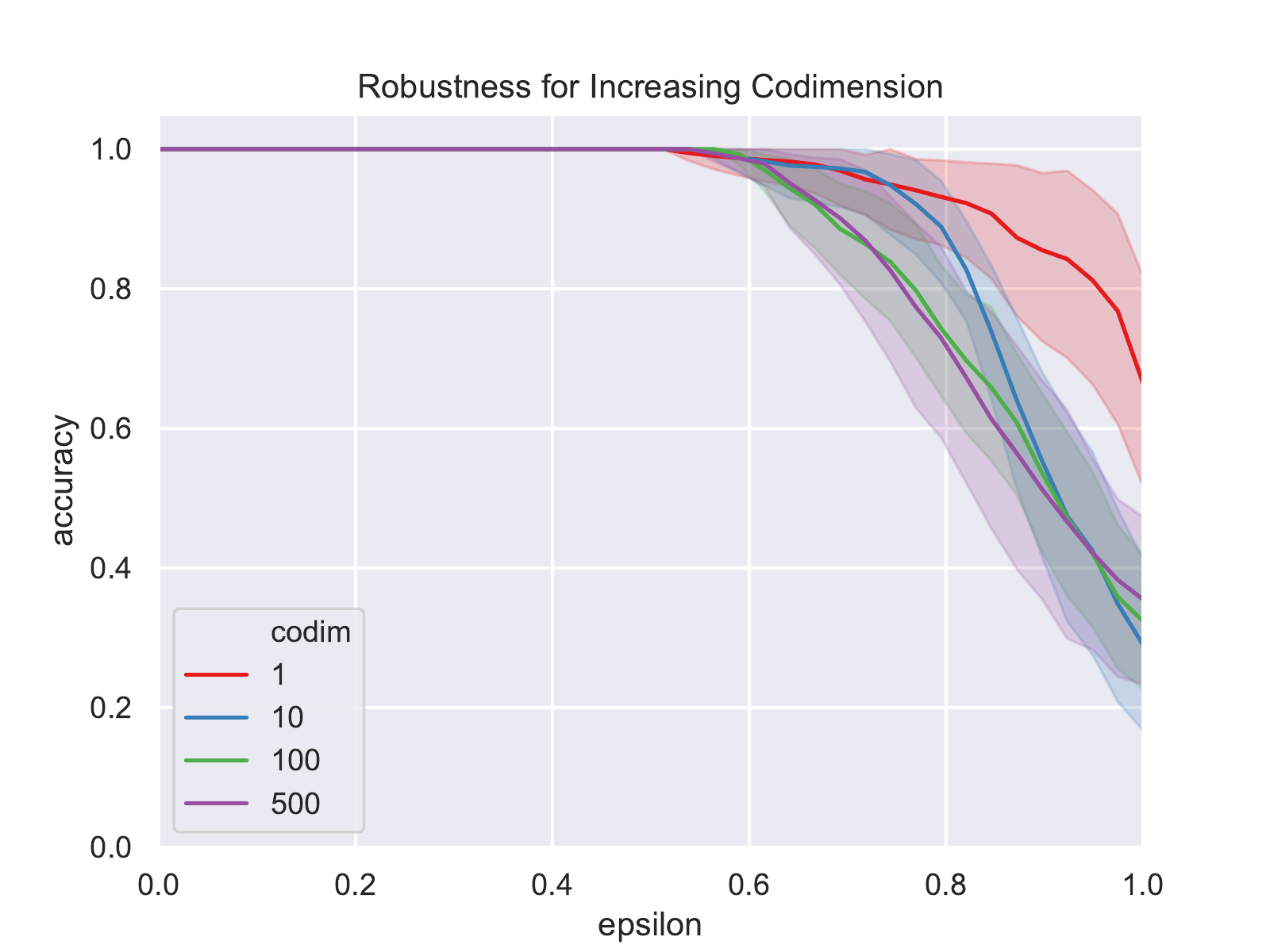}
\end{subfigure}
\begin{subfigure}{0.46\textwidth}
\includegraphics[width=0.99\linewidth]{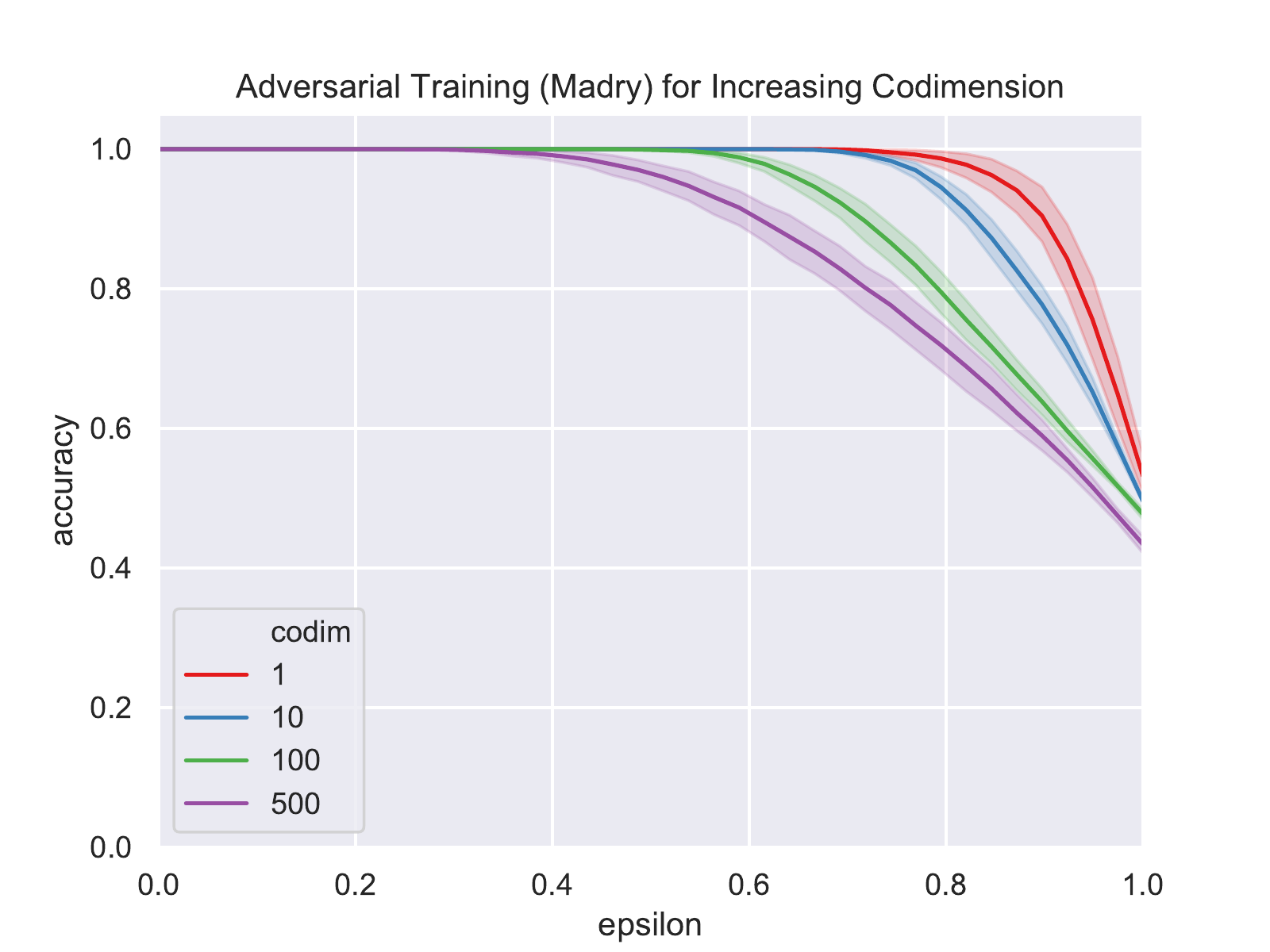}
\end{subfigure}
\begin{subfigure}{0.46\textwidth}
\includegraphics[width=0.99\linewidth]{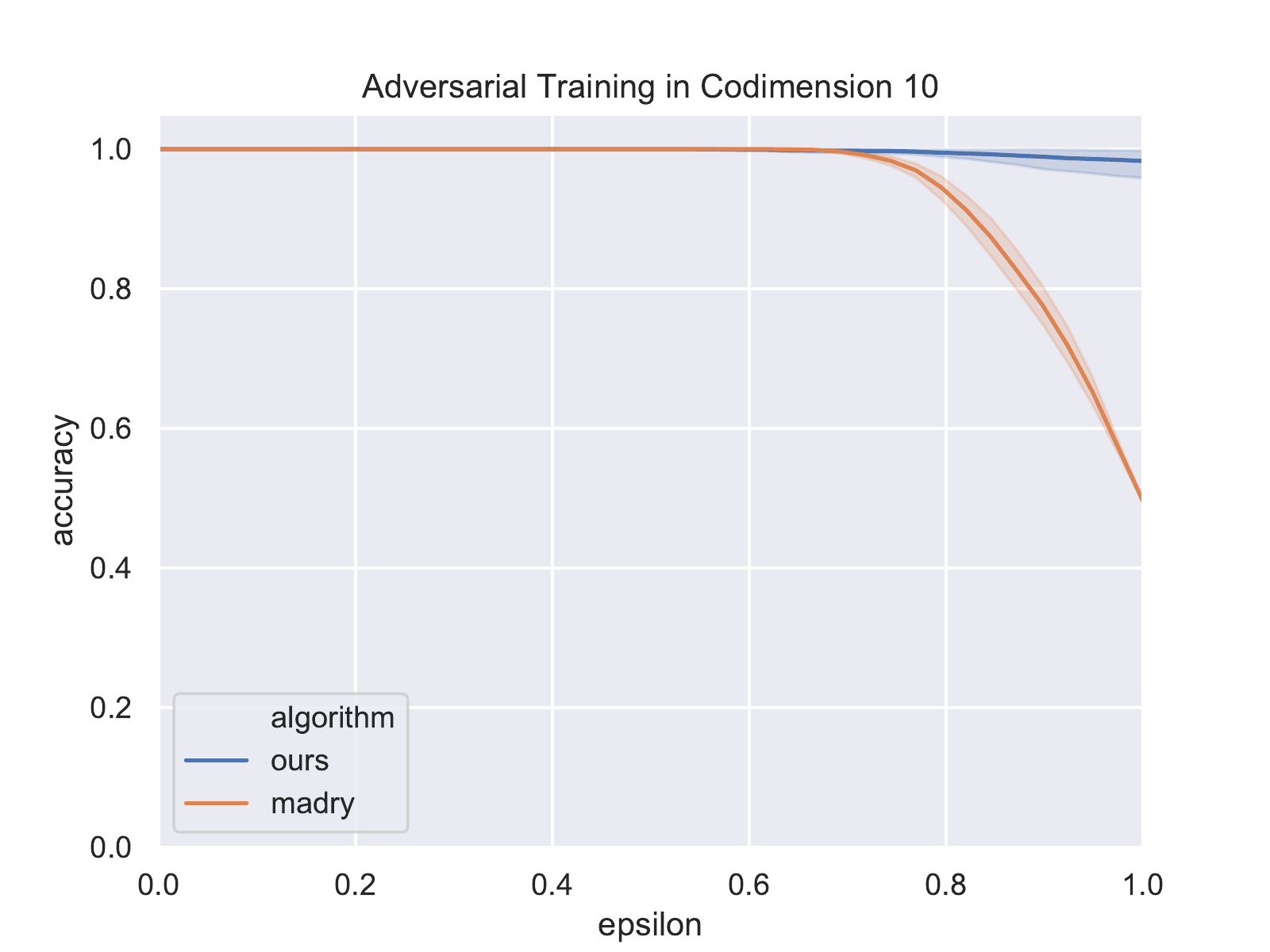}
\end{subfigure}
\begin{subfigure}{0.46\textwidth}
\includegraphics[width=0.99\linewidth]{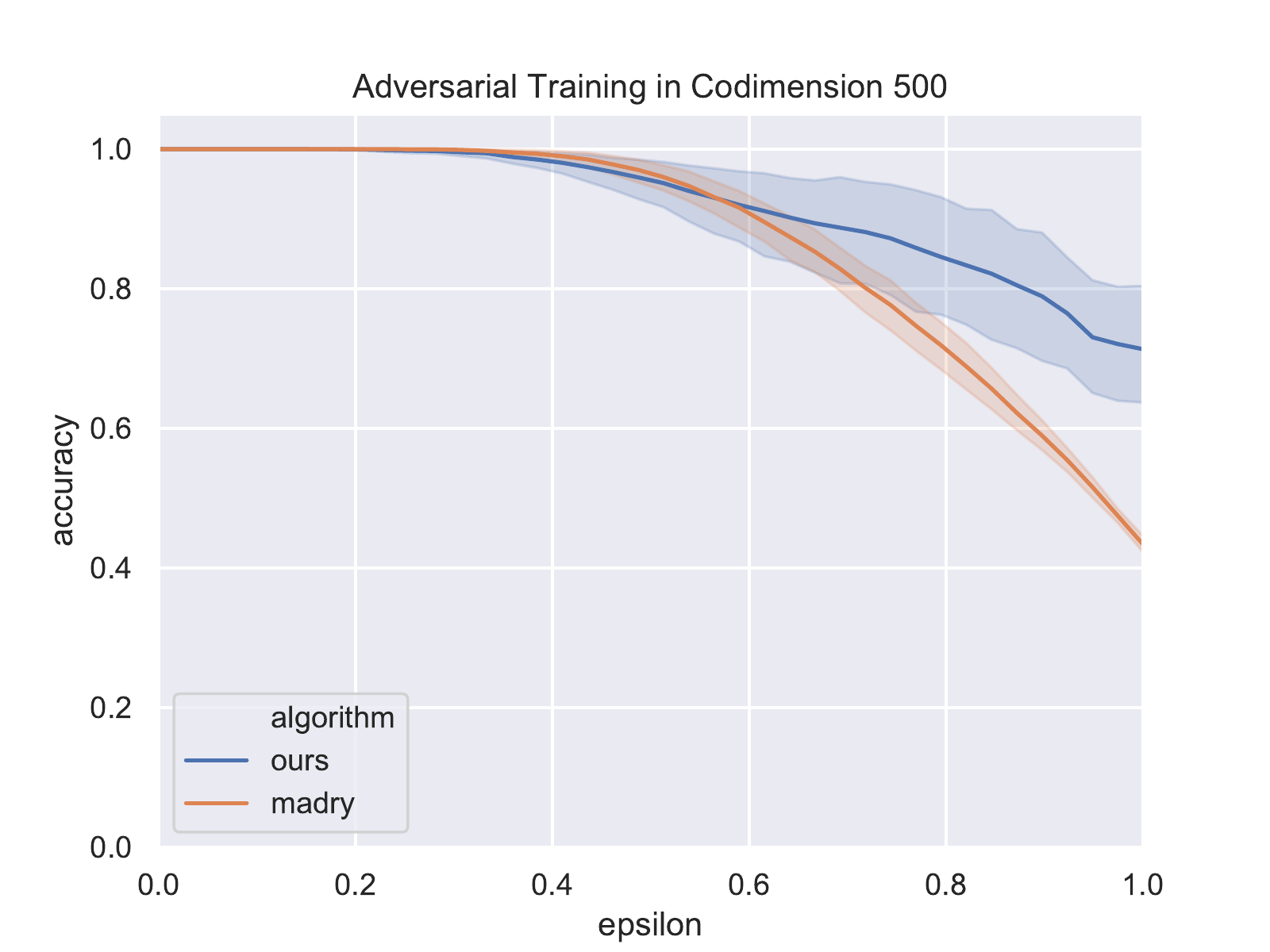}
\end{subfigure}
\caption{\textbf{Top Right:} As the codimension increases the robustness of decision boundaries learned by Adam on naturally trained networks for {\sc Circles} decreases steadily. \textbf{Top Left:} Training using the adversarial training procedure of \cite{Madry17} is no guarantee of robustness; as the codimension increases it becomes easier to find adversarial examples for {\sc Planes}. \textbf{Bottom:} Training using adversarial training with Voronoi constraints offers improved robustness in high codimension settings, on average.}
\label{fig:codimexp}
\end{center}
\end{figure}

Figure \ref{fig:codimexp} (Top Right, Bottom Left, Bottom Right) explores the use of adversarial training to improve robustness in high codimension settings for the {\sc Planes} dataset. As shown in Figure~\ref{fig:codimexp} (Top Right), as the codimension increases the adversarial training approach of \cite{Madry17} becomes less robust. This is because the $\|\cdot\|_{2}$ balls around $\Pi$ cover a smaller fraction of the tubular neighborhood around $\Pi$, as predicted by Theorem~\ref{thm:manifoldvolumelowerbound}. In Appendix~\ref{ssec:density} we show that even significantly increasing the sampling density does not notably improve robustness in high codimensions. 

Replacing the $\|\cdot\|_{2}$ ball constraint with the Voronoi cells improves robustness in high codimension settings, on average. In codimension 10 (Figure~\ref{fig:codimexp} (Bottom Left)), our approach achieves NAUC of $0.99$, while Madry's approach achieves NAUC of $0.94$. In codimension 500 (Figure~\ref{fig:codimexp} (Bottom Right)), our approach achieves NAUC of $0.92$, while Madry's approach achieves NAUC of $0.87$. 

\subsection{MNIST and CIFAR-10}
To explore the performances of adversarial training with Voronoi constraints on more realistic datasets, we evaluate on MNIST and CIFAR-10 and compare against the robust pretrained models of \cite{Madry17}.
%\footnote{\url{https://github.com/MadryLab/mnist_challenge}}\footnote{\url{https://github.com/MadryLab/cifar10_challenge}}. 

\begin{figure}[h!]
\begin{center}
\begin{subfigure}{0.47\textwidth}
\includegraphics[width=0.99\linewidth]{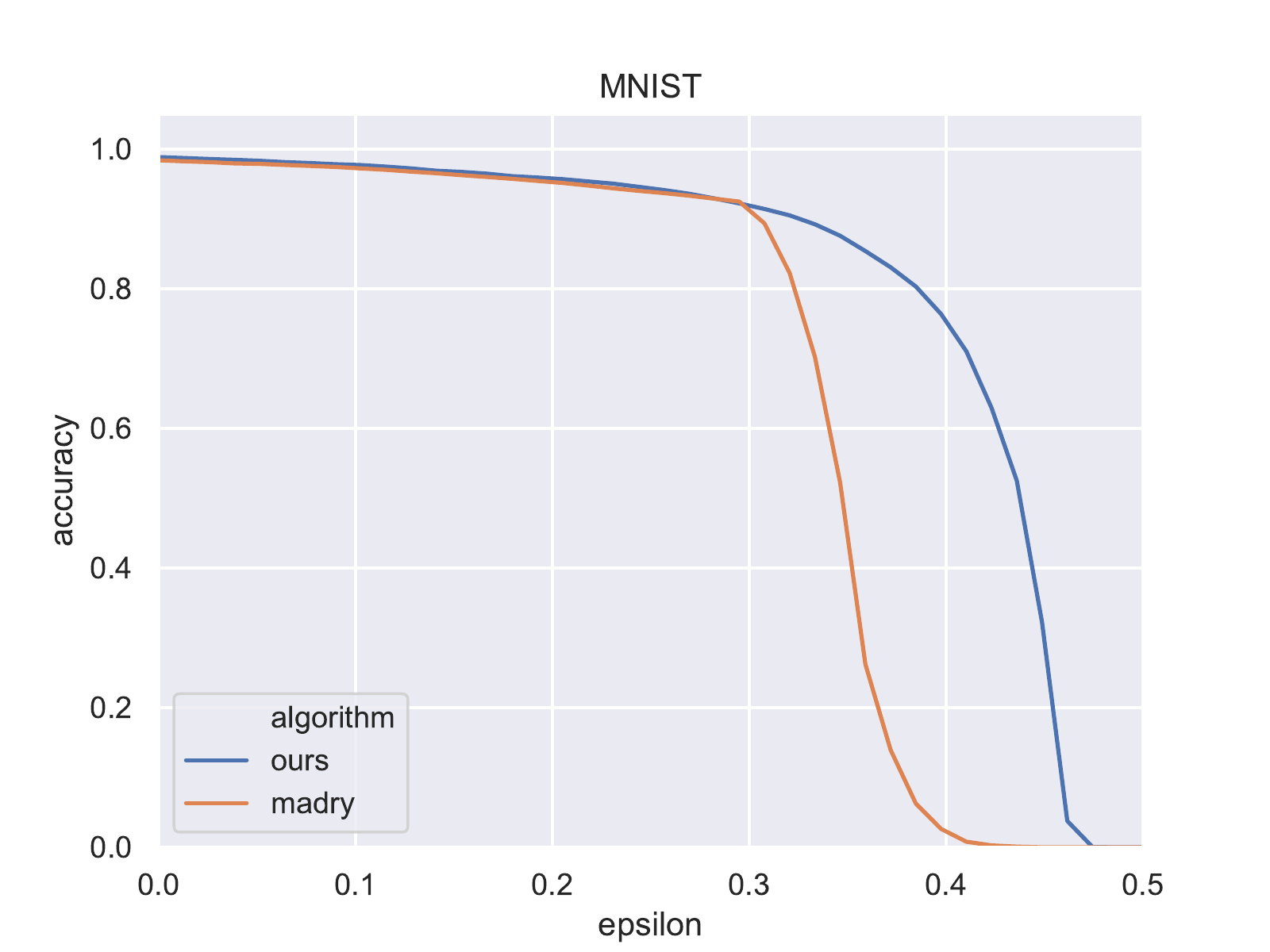}
\end{subfigure}
\begin{subfigure}{0.49\textwidth}
\includegraphics[width=0.99\linewidth]{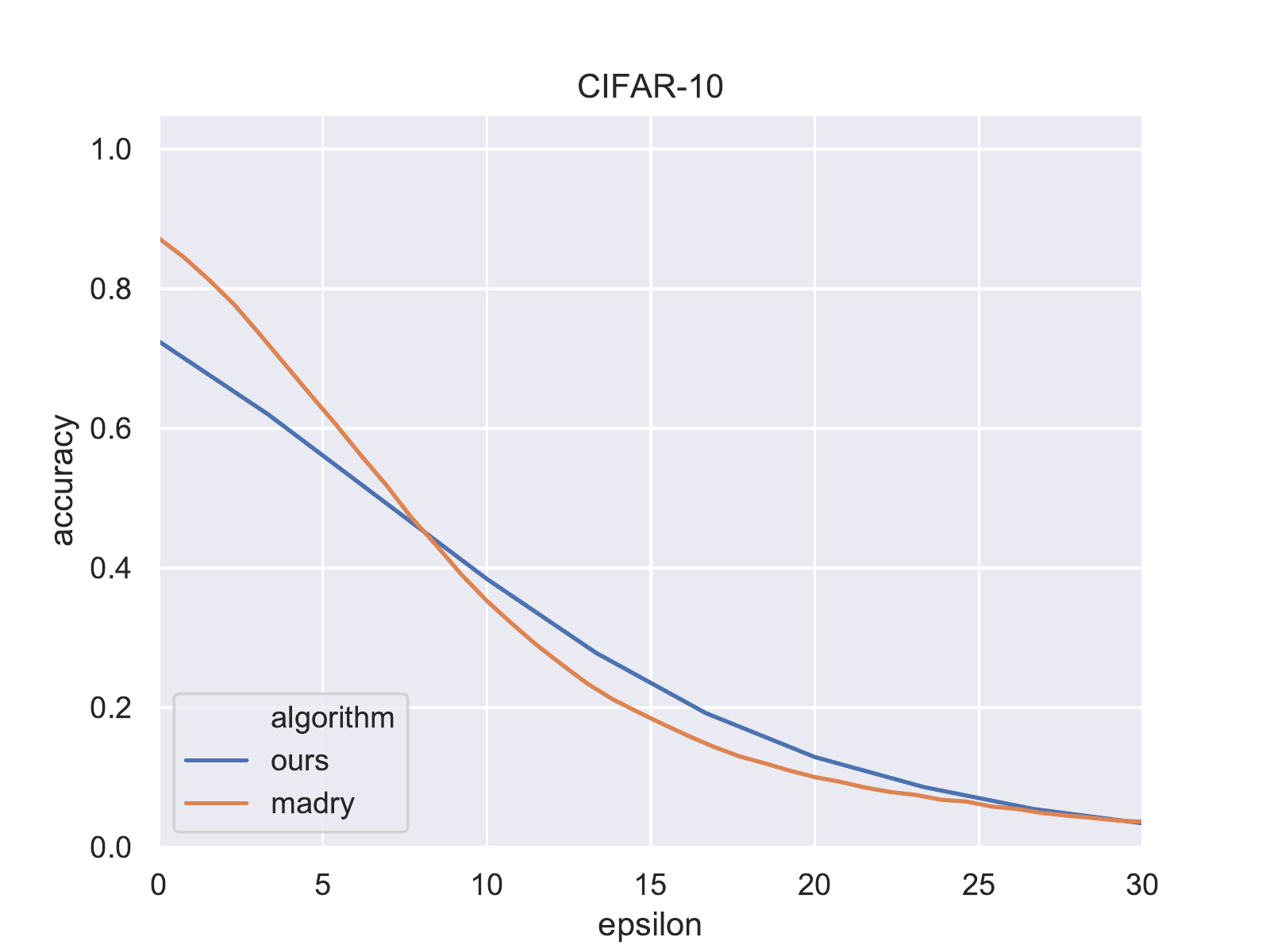}
\end{subfigure}
\caption{\textbf{Left:} Adversarial training with Voronoi constraints on MNIST. Our model has NAUC $0.81$ and high classification accuracy after $\epsilon=0.3$. In particular, our model maintains $76.3\%$ accuracy at $\epsilon=0.4$, compared to $2.6\%$ accuracy for the Madry model. \textbf{Right:} On CIFAR-10, both models achieve NAUC of $0.29$, but our model trades natural accuracy for robustness to larger perturbations.}
\label{fig:mnistcifar}
\end{center}
\end{figure}

Figure~\ref{fig:mnistcifar} (Left) shows that our model maintains near identical robustness to the Madry model on MNIST up to $\epsilon = 0.3$, after which our model \emph{significantly} outperforms the Madry model. The Madry model was explicitly trained for $\epsilon = 0.3$ perturbations. We emphasize that one advantage of our approach is that we did not need to set a value for the maximum perturbation size $\epsilon$. The Voronoi cells adapt to the maximum size allowable locally on the data distribution. Our model maintains $76.3\%$ accuracy at $\epsilon = 0.4$ compared to $2.6\%$ accuracy for the Madry model. Furthermore our model achieves NAUC of $0.81$, while the Madry model achieves NAUC of  $0.67$, an improvement of $20.8\%$ over the baseline. To our knowledge, this is the most robust MNIST model to $\|\cdot\|_{\infty}$ attacks.

Figure~\ref{fig:mnistcifar} (Right) shows the results of our approach on CIFAR-10. Both our model and the Madry model achieve NAUC of $0.29$. However our approach trades natural accuracy for increased robustness against larger perturbations. This tradeoff is well-known and explored in \cite{Tsipras19}. 

\begin{figure}[h!]
\begin{center}
\begin{subfigure}{0.5\textwidth}
\includegraphics[width=0.99\linewidth]{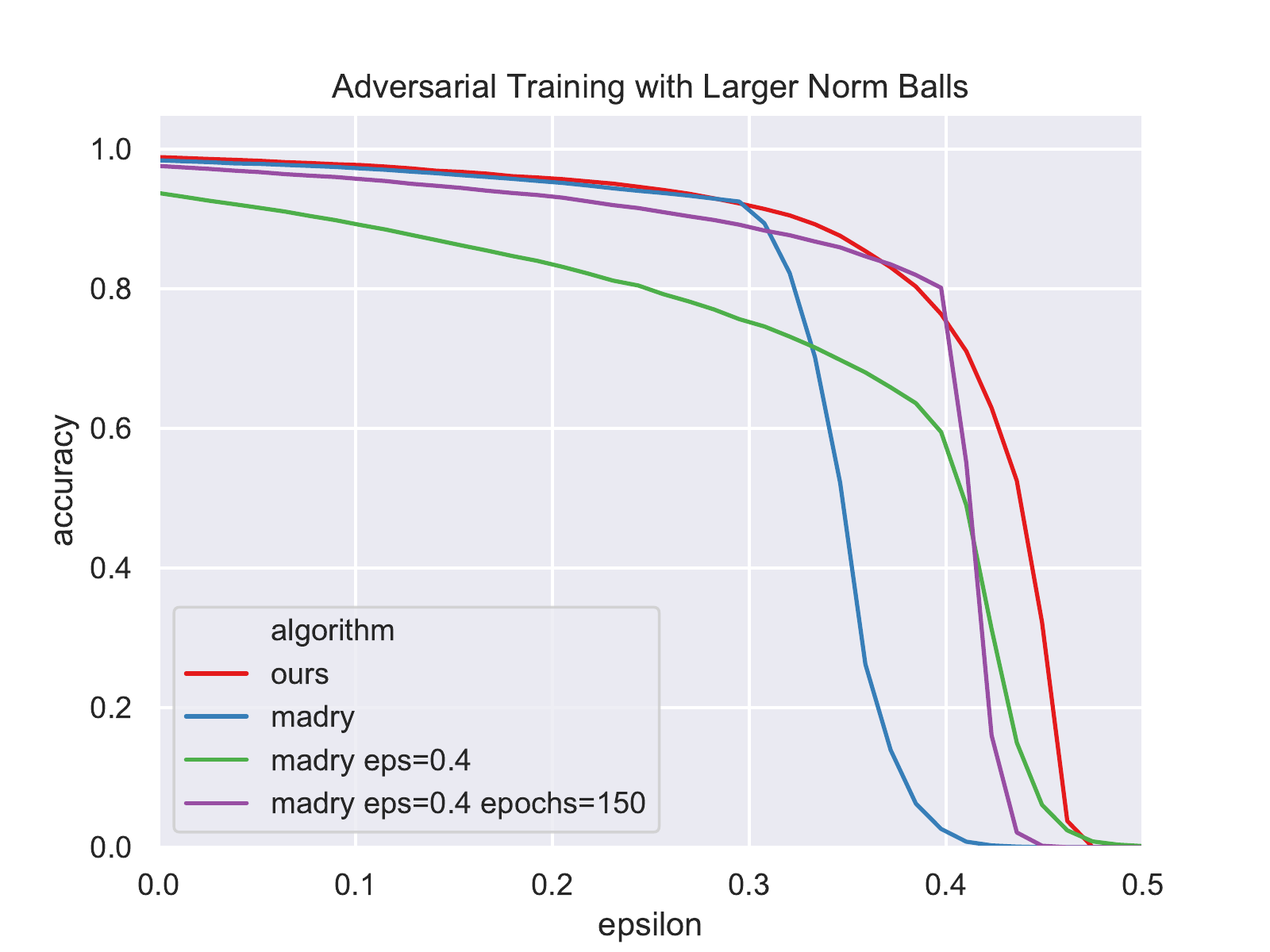}
\end{subfigure}
\caption{The adversarial training of \cite{Madry17} with $\epsilon = 0.4$ (shown in green) produces a model with significantly reduced robustness in the range $[0, 0.3]$. Increasing the number of epochs to $150$, the resulting model (shown in purple) does exhibit improved robustness in the range $[0.3, 0.4]$, at the expense of some robustness in the range $[0, 0.3]$ and still exhibits a sharp drop in accuracy after $0.4$. The purple model achieves NAUC of $0.76$, while our model achieves NAUC $0.81$.}
\label{fig:largernormball}
\end{center}
\end{figure}

A natural approach to improving the robustness of models produced by the adversarial training paradigm of \cite{Madry17} is to simply increase the maximum allowable perturbation size $\epsilon$ of the norm ball constraint. As shown in Figure~\ref{fig:largernormball}, increasing the size of $\epsilon$ to $0.4$, from the $0.3$ with which \cite{Madry17} originally trained, and training for only $100$ epochs produces a model which exhibits significantly worse robustness in the range $[0, 0.3]$ than the pretrained model. If we increase the number of training epochs to $150$, the approach of \cite{Madry17} with $\epsilon = 0.4$ produces a model with improved robustness in the range $[0.3, 0.4]$, but that still exhibits the sharp drop in accuracy after $0.4$. Additionally the model trained with $\epsilon=0.4$ for $150$ epochs performs worse than both the pretrained model and our model in the range $[0, 0.3]$. Our model achieves NAUC $0.81$, while the model trained with $\epsilon = 0.4$ for $150$ epochs achieves NAUC $0.76$. We emphasize that our approach does not require us to set $\epsilon$, which is particularly important in practice where the maximum amount of robustness achievable may not be known a-priori. 

\section{Conclusions}
The $\|\cdot\|_{p}$-ball constraint for describing adversarial perturbations has been a productive formalization for designing robust deep networks. However, the use of $\|\cdot\|_{p}$-balls has significant drawbacks in high-codimension settings and leads to sub-optimal results in practice. Adversarial training with Voronoi constraints improves robustness by giving the adversary the freedom to explore $\mathcal{M}^{\epsilon}$ and generate adversarial examples close to $\Lambda_p$.  

\bibliography{geomae}

\appendix

\section{Omitted Proofs}
\subsection{Proof of Theorem 2}
\begin{figure}
\begin{center}
\includegraphics[width=0.5\linewidth]{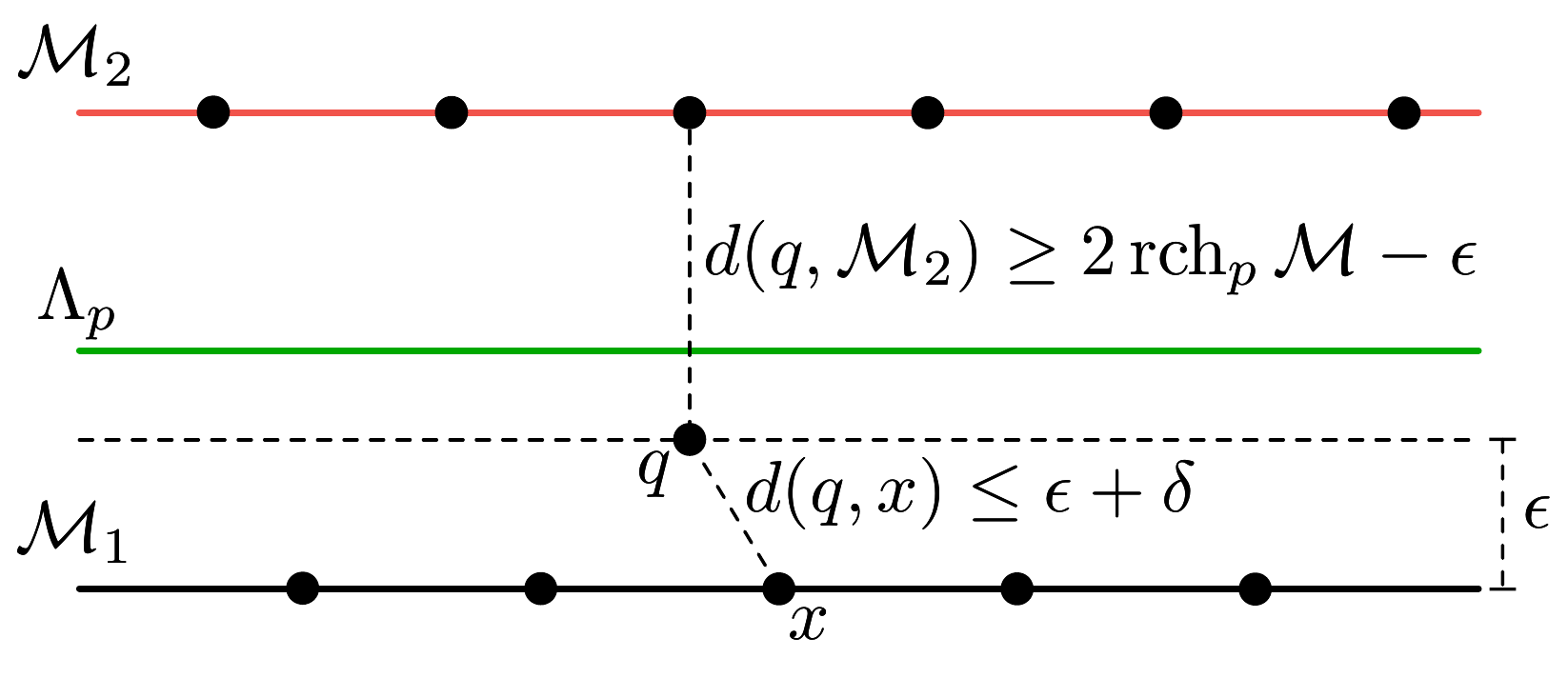}
\caption{Proof of Theorem 2. The distance from a query point $q$ to $\mathcal{M}_{2}$, and thus the closest incorrectly labeled sample, is lower bounded by the distance necessary to reach the medial axis $\Lambda_{p}$ plus the distance from $\Lambda_{p}$ to $\mathcal{M}_{2}$.} 
\label{fig:samplingproof}
\end{center}
\end{figure}

\begin{proof}
Here we use $d(\cdot, \cdot)$ to denote the metric induced by the $\|\cdot\|_{p}$ norm. We begin by proving (1). Let $q \in \mathcal{M}^{\epsilon}$ be any point in $\mathcal{M}^{\epsilon}$. Suppose without loss of generality that $q \in \mathcal{M}_{i}^{\epsilon}$ for some class $i$. The distance $d(q, \mathcal{M}_{j})$ from $q$ to any other data manifold $\mathcal{M}_{j}$, and thus any sample on $\mathcal{M}_{j}$, is lower bounded by $d(q,\mathcal{M}_{j}) \geq 2\rch_p{\Lambda_p} - \epsilon$. See Figure \ref{fig:samplingproof}. It is then both necessary and sufficient that there exists a $x \in \mathcal{M}_{i}$ such that $d(q, x) < 2\rch_p{\Lambda_p} - \epsilon$ for $f_{\nn}(q) = i$. (Necessary since a properly placed sample on $\mathcal{M}_{j}$ can achieve the lower bound on $d(q, \mathcal{M}_{j})$.) The distance from $q$ to the nearest sample $x$ on $\mathcal{M}_{i}$ is $d(q, x) \leq \epsilon + \delta$ for some $\delta > 0$. The question is how large can we allow $\delta$ to be and still guarantee that $f_{\nn}$ correctly classifies $\mathcal{M}^{\epsilon}$? We need
\begin{equation*}
d(q, x) \leq \epsilon + \delta \leq 2\rch_p{\Lambda_p} - \epsilon \leq d(q, \mathcal{M}_{j})
\end{equation*}
which implies that $\delta \leq 2 (\rch_p{\Lambda_p} - \epsilon)$. It follows that a $\delta$-cover with $\delta = 2(\rch_p{\Lambda_p} - \epsilon)$ is sufficient, and in some cases necessary, to guarantee that $f_{nn}$ correctly classifies $\mathcal{M}^{\epsilon}$.

Next we prove (2). As before let $q \in \mathcal{M}_{i}^{\epsilon}$. It is both necessary and sufficient for $q \in B(x, \rch_p{\Lambda_p})$ for some sample $x \in \mathcal{M}_{i}$ to guarantee that $f_{\mathcal{L}}(q) = i$, by definition of $\mathcal{L}$. The distance to the nearest sample $x$ on $\mathcal{M}_{i}$ is $d(q, x) \leq \epsilon + \delta$ for some $\delta > 0$. Thus it suffices that $\delta \leq \rch_p{\Lambda_p}-\epsilon$. 
\end{proof}

\subsection{Proof of Theorem 3}
\begin{proof}
Let $q \in \Pi_1^{\epsilon}$. Since $\Pi_1$ is flat, the distance from $q$ to the nearest sample $x \in \Pi_1$ is bounded as $\|q - x\|_{2} \leq \sqrt{\epsilon^2 + \delta^2}$. For $f_{\nn}(q) = 1$ we need that $\|q - x\|_{2} \leq 2 - \epsilon$, and so it suffices that $\delta \leq 2 \sqrt{1 - \epsilon}$. In this setting, this is also necessary; should $\delta$ be any larger a property placed sample on $\Pi_{2}$ can claim $q$ in its Voronoi cell.

Similarly for $f_{\mathcal{L}}(q) = 1$ we need that $\|q - x\|_{2} \leq 1$, and so it suffices that $\delta \leq \sqrt{1-\epsilon^2}$. In this setting, this is also necessary; should $\delta$ be any larger, $q$ lies outside of every $\|\cdot\|_{2}$-ball $B(x, 1)$ and so $\mathcal{L}$ is free to learn a decision boundary that misclassifies $q$.

Let $\mathcal{N}(\delta, \mathcal{M})$ denote the size of the minimum $\delta$-cover of $\mathcal{M}$. Since $\Pi$ is flat (has no curvature) and since the intersection of $\Pi$ with a $d$-ball centered at a point on $\Pi$ is a $k$-ball, a standard volume argument can be applied in the affine subspace $\operatorname{aff}{\Pi}$ to conclude that $\mathcal{N}(\delta,\Pi) \in \Theta\left(\vol_{k}\Pi/\delta^{k}\right)$. So we have
\begin{align*}
\frac{\mathcal{N}(\sqrt{1-\epsilon^2}, \Pi)}{\mathcal{N}(2\sqrt{1-\epsilon}, \Pi)} &\in \Omega\left(2^{k} \left(\frac{1}{1 + \epsilon}\right)^{k/2}\right)\\
&\in \Omega\left(2^{k /2}\right)
\end{align*}
\end{proof}

\subsection{Proof of Theorem 4}
\begin{proof}
Assuming the balls centered on the samples in $X$ are disjoint we get the upper bound 
\begin{equation}
\label{equ:manifoldupperbound}
\vol{X^{\epsilon}} \leq \vol{B_{\epsilon} |X|} = \frac{\pi^{d/2}}{\Gamma(\frac{d}{2}+1)}\epsilon^d |X|.
\end{equation}

The medial axis $\Xi$ of $\mathcal{M}$ is defined as the closure of the set of all points in $\R^d$ that have two or more closest points on $\mathcal{M}$ in the norm $\|\cdot\|_{2}$. The medial axis $\Xi$ is similar to the decision axis $\Lambda_2$, except that the nearest points do not need to be on distinct class manifolds. For $\epsilon \leq \rch_{2}{\Xi}$, we have the lower bound

\begin{equation}
\label{equ:manifoldtubelowerbound}
\vol{\mathcal{M}^{\epsilon}} \geq \vol_{d-k}{B^{d-k}_{\epsilon}} \vol_{k}{\mathcal{M}} = \frac{\pi^{(d-k)/2}}{\Gamma\left(\frac{d-k}{2}+1\right)}\epsilon^{d-k}  \vol_{k}{\mathcal{M}}.
\end{equation}

Combining Equations~\ref{equ:manifoldupperbound} and \ref{equ:manifoldtubelowerbound} gives the result. To get the asymptotic result we apply Stirling's approximation to get
\begin{align*}
\frac{\Gamma(\frac{d-k}{2}+1)}{\Gamma(\frac{d}{2}+1)} &\approx (2e)^{k/2} \frac{(d - k)^{(d - k + 1) / 2}}{d^{(d+1)/2}}\\
                                                      &= (2e)^{k/2} \frac{\left(\frac{d - k}{d}\right)^{(d+1)/2}}{(d - k)^{k/2}}\\
                                                      &= (2e)^{k/2} \frac{\left(1 - \frac{k}{d}\right)^{(d+1)/2}}{(d - k)^{k/2}}\\ 
                                                      &\approx \left(\frac{2}{d - k}\right)^{k/2}.
\end{align*}
The last step follows from the fact that $\lim_{d \rightarrow \infty} (1 - k/d)^{(d+1)/2} = e^{-k/2}$, where $e$ is the base of the natural logarithm.
\end{proof}

\section{Additional Experiments}

\subsection{Increasing Sampling Density}
\label{ssec:density}
\begin{figure}[h!]
\begin{center}
\begin{subfigure}{0.32\textwidth}
\includegraphics[width=0.99\linewidth]{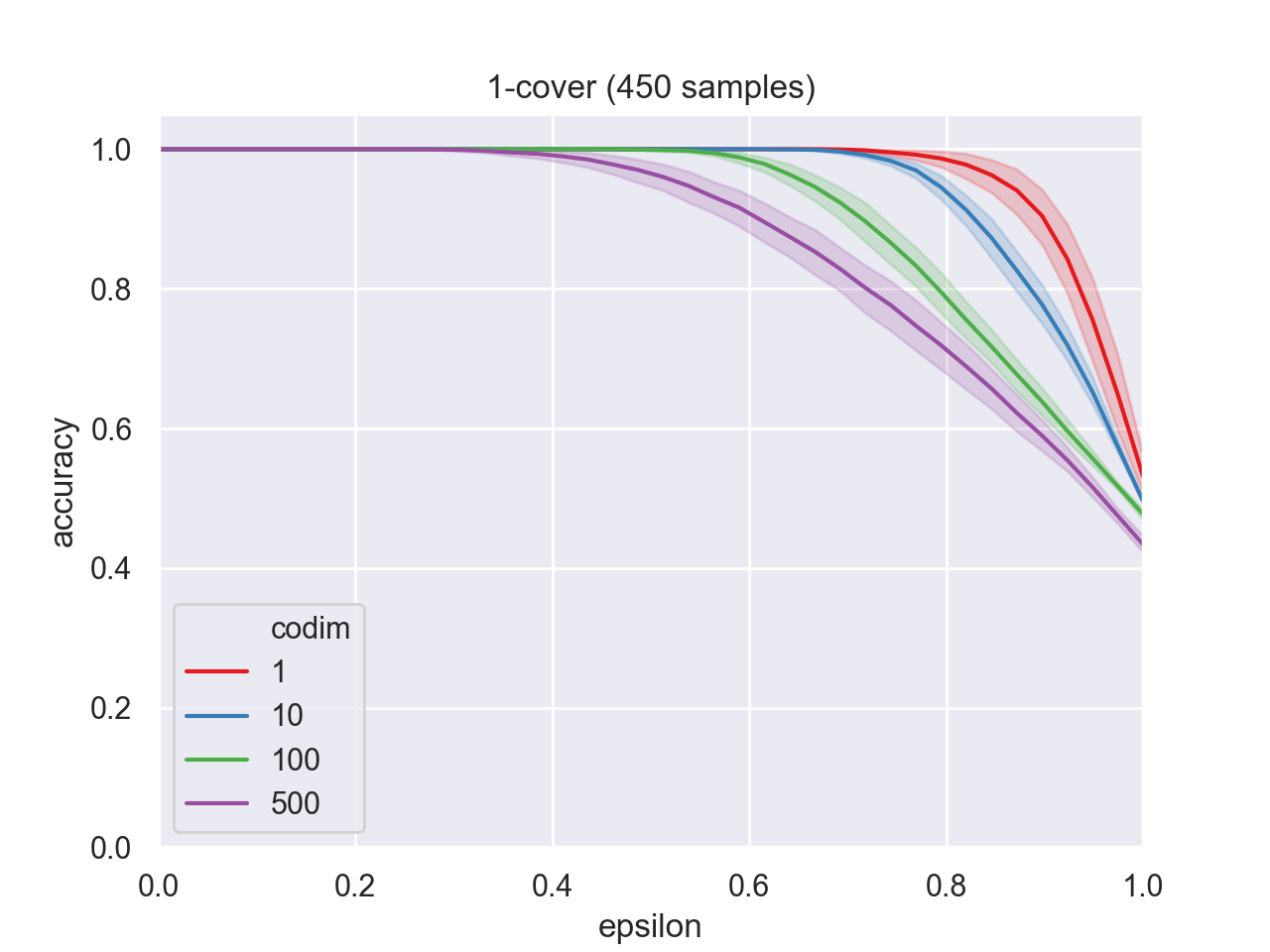}
\end{subfigure}
\begin{subfigure}{0.32\textwidth}
\includegraphics[width=0.99\linewidth]{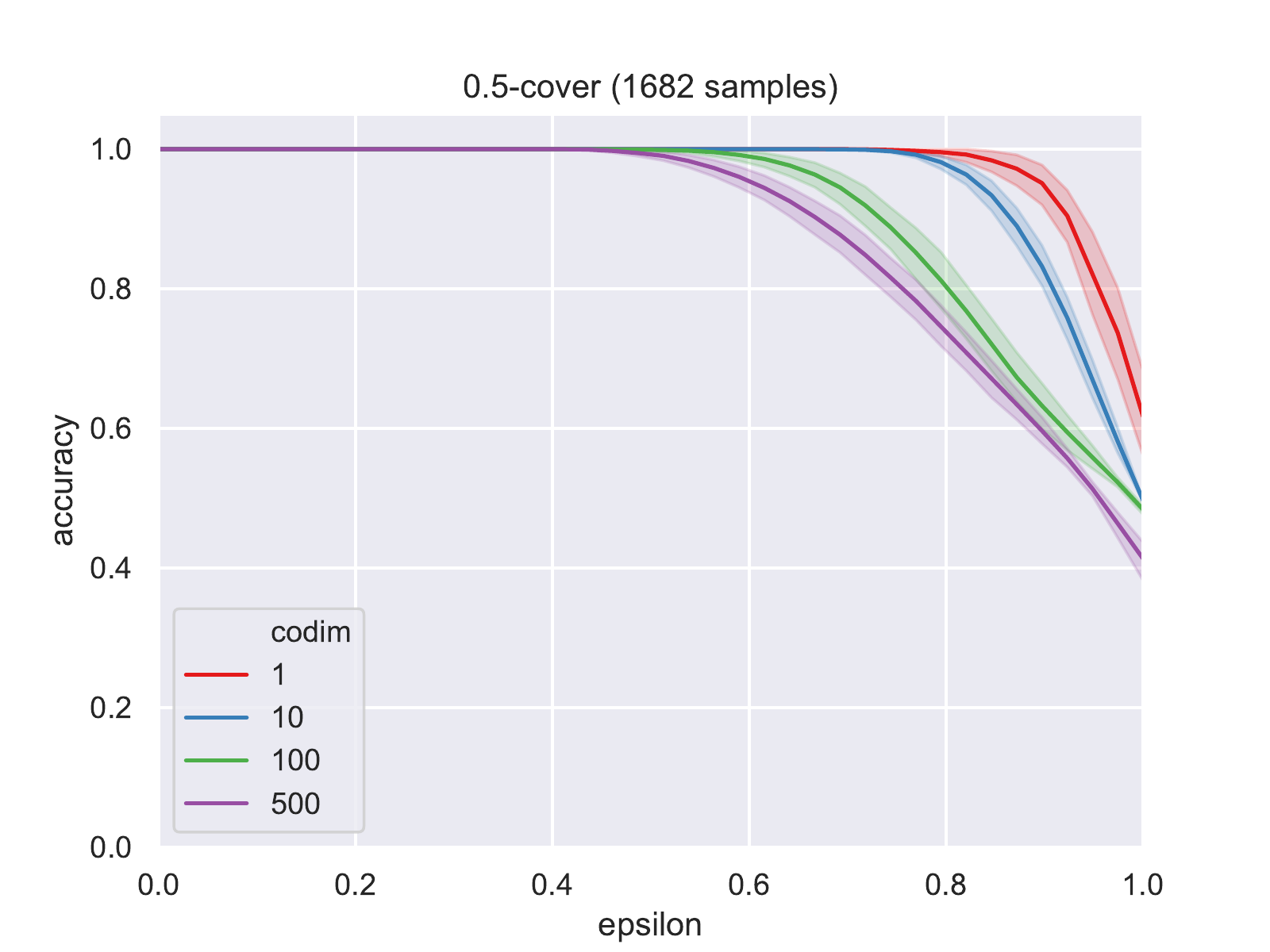}
\end{subfigure}
\begin{subfigure}{0.32\textwidth}
\includegraphics[width=0.99\linewidth]{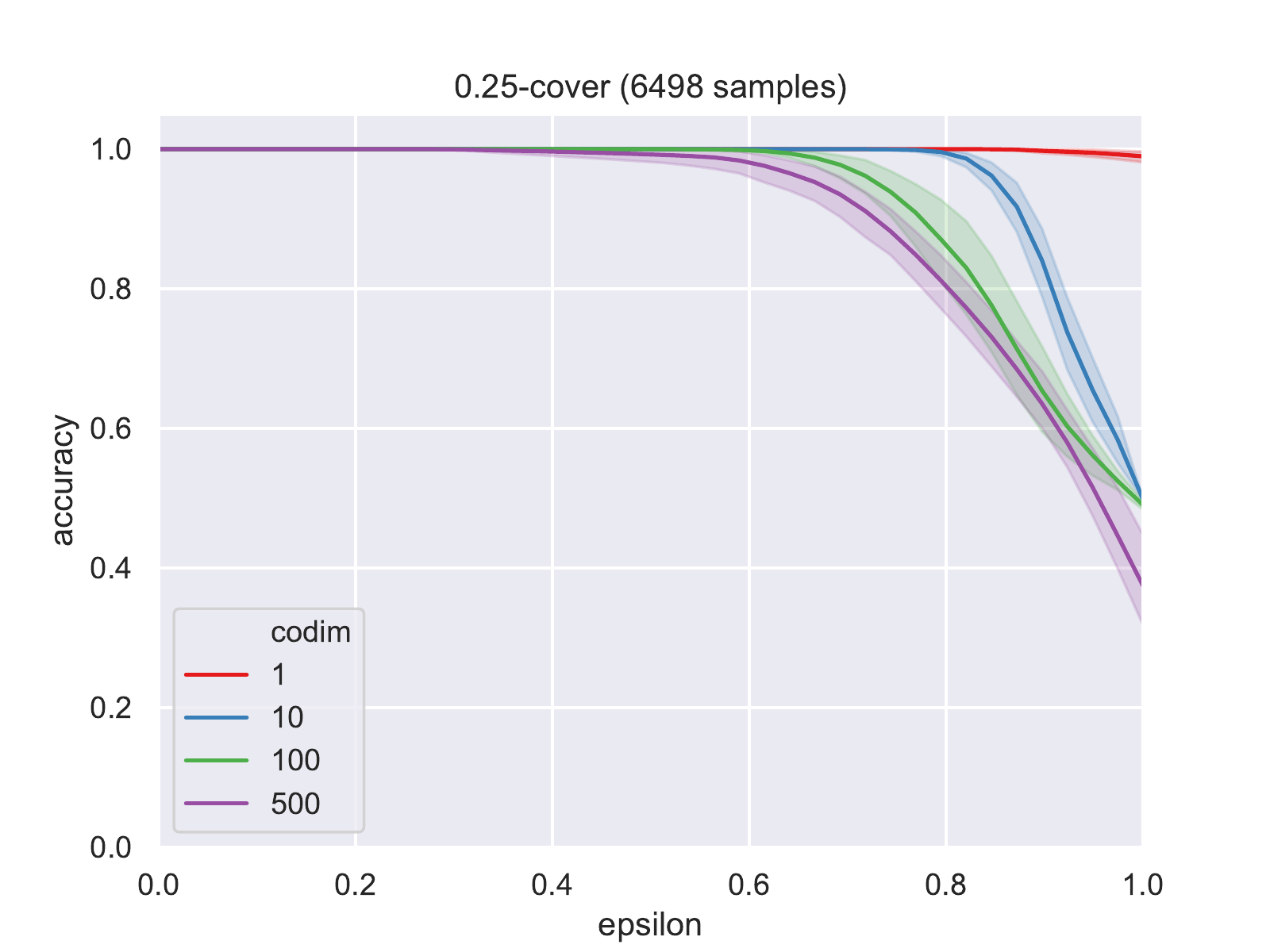}
\end{subfigure}
\caption{Adversarial training of \cite{Madry17} on the {\sc Planes} dataset with a $1$-cover (left), consisting of $450$ samples, a $0.5$-cover (center), $1682$ samples, and a $0.25$-cover (right), $6498$ samples. Increasing the sampling density improves robustness at the same codimension, but the improvement is much less notable in high-codimension.}
\label{fig:samplingdensityexp}
\end{center}
\end{figure}

The {\sc Planes} dataset is sampled so that the trianing set is a $1$-cover of the underlying planes, which requires 450 sample points. Figure~\ref{fig:samplingdensityexp} shows the results of increasing the sampling density to a $0.5$-cover (1682 samples) and a $0.25$-cover (6498 samples). In low-codimension, increasing the sampling density improves the robustness of adversarial training. However, in high-codimension, even a substantial increase in the number of samples gives a only a small improvement in robustness.

\end{document}